\documentclass[graybox]{svmult}

\usepackage{mathptmx}       % selects Times Roman as basic font
\usepackage{helvet}         % selects Helvetica as sans-serif font
\usepackage{courier}        % selects Courier as typewriter font
\usepackage{type1cm}        % activate if the above 3 fonts are
\usepackage{makeidx}         % allows index generation
\usepackage{graphicx}        % standard LaTeX graphics tool
\usepackage{multicol}        % used for the two-column index
\usepackage[bottom]{footmisc}% places footnotes at page bottom

\makeindex             % used for the subject index
\begin{document}

\title*{Identifying and Harnessing the Building Blocks of Machine Learning Pipelines for Sensible Initialization of a Data Science Automation Tool}
\titlerunning{Building Blocks for Sensible Initialization of a Data Science Automation Tool}
\author{Randal S.~Olson and Jason H.~Moore}
\institute{Randal S.~Olson \at Institute for Biomedical Informatics, University of Pennsylvania, \email{rso@randalolson.com}
\and Jason H.~Moore \at Institute for Biomedical Informatics, University of Pennsylvania, \email{jhmoore@upenn.edu}}
\maketitle

\abstract{As data science continues to grow in popularity, there will be an increasing need to make data science tools more scalable, flexible, and accessible. In particular, automated machine learning (AutoML) systems seek to automate the process of designing and optimizing machine learning pipelines. In this chapter, we present a genetic programming-based AutoML system called TPOT that optimizes a series of feature preprocessors and machine learning models with the goal of maximizing classification accuracy on a supervised classification problem. Further, we analyze a large database of pipelines that were previously used to solve various supervised classification problems and identify 100 short series of machine learning operations that appear the most frequently, which we call the {\em building blocks} of machine learning pipelines. We harness these building blocks to initialize TPOT with promising solutions, and find that this sensible initialization method significantly improves TPOT's performance on one benchmark at no cost of significantly degrading performance on the others. Thus, sensible initialization with machine learning pipeline building blocks shows promise for GP-based AutoML systems, and should be further refined in future work.}


\section{Introduction}
\label{sec:introduction}

\begin{figure}[b]
\sidecaption
\includegraphics[width=\textwidth]{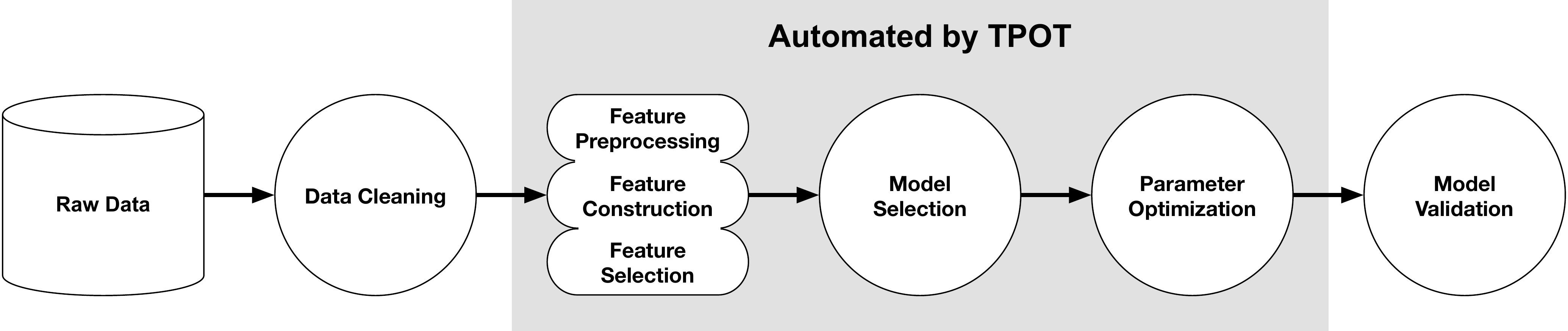}
\caption{A typical machine learning pipeline. Machine learning practitioners often start with a raw data set that must be formatted, have missing values imputed, and otherwise prepared for analysis. Following this step, practitioners must often transform the feature set into a format that is amenable to modeling, for example by preprocessing the features via scaling, constructing new features from existing features, or removing less useful features via feature selection. Next, practitioners must select a machine learning model to fit to the data, then optimize the parameters of the model and feature transformation operations to allow the model to best capture the underlying signal in the data. At the end of this process, practitioners must evaluate their pipeline on a validation data set that the pipeline never saw before, which allows the practitioners to determine whether the pipeline generalizes beyond the initial training data.}
\label{fig:tpot-ml-pipeline-diagram}
\end{figure}

Machine learning is often touted as a ``field of study that gives computers the ability to learn without being explicitly programmed''~\cite{Simon2013}. Despite this common claim, it is well-known by machine learning practitioners that designing effective machine learning pipelines is often a tedious endeavor, and typically requires considerable experience with machine learning algorithms, expert knowledge of the problem domain, and brute force search to accomplish. Figure~\ref{fig:tpot-ml-pipeline-diagram} depicts a typical machine learning pipeline, where each step requires intervention by machine learning practitioners. Thus, contrary to what machine learning enthusiasts would have us believe, machine learning still requires considerable explicit programming.

In response to this challenge, several automated machine learning methods have been developed over the years~\cite{Hutter2015}. Over the past year, we have been developing a Tree-based Pipeline Optimization Tool (TPOT) that automatically designs and optimizes machine learning pipelines for a given problem domain~\cite{Olson2016EvoBio}, without any need for human intervention. In short, TPOT optimizes machine learning pipelines using a version of genetic programming (GP), a well-known evolutionary computation technique for automatically constructing computer programs~\cite{Koza1992,Banzhaf1998}. Previously, we demonstrated that combining GP with Pareto optimization enables TPOT to automatically construct high-accuracy {\em and} compact pipelines that consistently outperform basic machine learning analyses~\cite{Olson2016GECCO}. In this chapter, we report on our progress toward introducing sensible initialization~\cite{Greene2009} of the GP population into TPOT, with the goal of enabling TPOT to harness expert knowledge about machine learning pipelines to efficiently discover effective pipelines for a given problem domain.

\newpage

Previous research on the initialization of GP populations has shown that the initialization process can vitally affect the performance of GP algorithms~\cite{Luke01,ONeill2003,GarciaArnau2007}. However, most of this research has focused on generating a diversity of valid GP tree structures, which may not be useful in all application domains. Here we follow in the footsteps of~\cite{Greene2009} and focus on harnessing expert knowledge---in this case, expert knowledge about machine learning pipelines---to initialize the GP population. In particular, we attempt to identify the building blocks~\cite{Goldberg2002} of machine learning pipelines, and harness these building blocks for sensible initialization of the GP population in TPOT.

\section{Automated Machine Learning}
\label{sec:automl}

In the early days of machine learning automation research (AutoML for short), researchers focused primarily on hyperparameter optimization~\cite{Hutter2015}. For example, the most commonly-used form of hyperparameter optimization is {\em grid search}, where users apply brute force search to evaluate a predefined range of model parameters to find the model parameters that allows for the best model fit. More recently, researchers showed that it is possible to discover optimal parameter sets faster than exhaustive grid search by randomly sampling within a predefined grid search~\cite{Bergstra2012}, which shows promise for guided search in the hyperparameter space. Bayesian optimization, in particular, has proven effective for hyperparameter optimization and has even outperformed manual hyperparameter tuning by expert practitioners~\cite{Snoek2012}.

Another focus of AutoML research has been feature construction. One recent example of automated feature construction is the ``Data Science Machine,'' which automatically constructs features from relational databases via deep feature synthesis. In their work,~\cite{Kanter2015} demonstrated the crucial role of automated feature construction in machine learning pipelines by entering their Data Science Machine in three machine learning competitions and achieving expert-level performance in all of them. Thus, we know that automated feature construction can play a vital role in AutoML systems.

More recently,~\cite{Fuerer2015} developed an AutoML system called {\em auto-sklearn}, which uses Bayesian optimization to discover the ideal combination of data and feature preprocessors, models, and model hyperparameters to maximize classification accuracy for a particular problem domain. However, auto-sklearn optimizes over a predefined set of pipelines that only include one data preprocessor, one feature preprocessor, and one model, which precludes auto-sklearn from producing arbitrarily large pipelines that may be important for AutoML.

~\cite{Zutty2015} demonstrated an AutoML system using genetic programming (GP) to optimize machine learning pipelines for signal processing, and found that GP is capable of designing better pipelines than humans for one signal processing task. As such, GP shows considerable promise in the AutoML domain, and we significantly extend this work with TPOT.

\section{Methods}
\label{sec:methods}

In the following sections, we provide an overview of the Tree-based Pipeline Optimization Tool (TPOT), including the machine learning operators used as genetic programming (GP) primitives, the tree-based pipelines used to combine the primitives into working machine learning pipelines, and the GP algorithm used to evolve said tree-based pipelines. We further describe the process we implemented to provide sensible initialization for the GP algorithm in TPOT, and conclude with a description of the data sets used to evaluate this new version of TPOT. TPOT is an open source project on GitHub, and the underlying Python code can be found at~\url{https://github.com/rhiever/tpot}.

\subsection{Machine Learning Pipeline Operators}
\label{sec:ml-pipeline-operators}

At its core, TPOT is a wrapper for the Python machine learning package, scikit-learn~\cite{scikit-learn}. Thus, each machine learning pipeline operator (i.e., GP primitive) in TPOT corresponds to a machine learning algorithm, such as a supervised classification model. All implementations of the machine learning algorithms listed below are from scikit-learn (except XGBoost), and we refer to the scikit-learn documentation~\cite{scikit-learn} and~\cite{MachineLearningBook} for detailed explanations of the machine learning algorithms used in TPOT.

{\bf Supervised Classification Operators.} DecisionTree, RandomForest, eXtreme Gradient Boosting Classifier (from XGBoost,~\cite{Chen2016}), LogisticRegression, and KNearestNeighborClassifier. Classification operators store the classifier's predictions as a new feature as well as the classification for the pipeline.

{\bf Feature Preprocessing Operators.} StandardScaler, RobustScaler, MinMaxScaler, MaxAbsScaler, RandomizedPCA~\cite{Martinsson2011}, Binarizer, and PolynomialFeatures. Preprocessing operators modify the data set in some way and return the modified data set.

{\bf Feature Selection Operators.} VarianceThreshold, SelectKBest, SelectPercentile, SelectFwe, and Recursive Feature Elimination (RFE). Feature selection operators reduce the number of features in the data set using some criteria and return the modified data set.

\begin{figure}[t]
\sidecaption
\includegraphics[width=\textwidth]{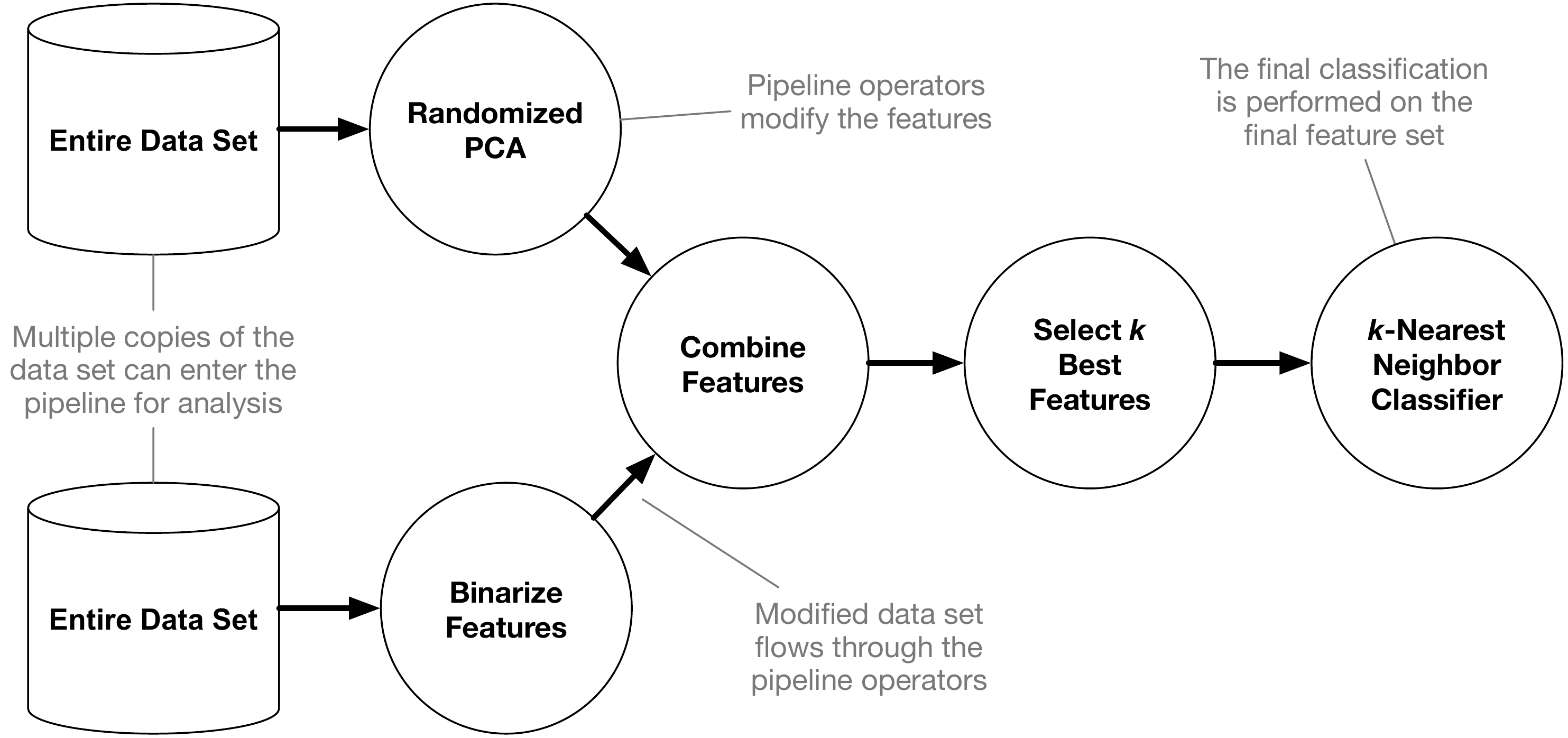}
\caption{An example tree-based pipeline from TPOT. Each circle corresponds to a machine learning operator, and the arrows indicate the direction of the data flow.}
\label{fig:tpot-pipeline-example}
\end{figure}

We also include an operator that combines disparate data sets, as demonstrated in Figure~\ref{fig:tpot-pipeline-example}, which allows multiple modified copies of the data set to be combined into a single data set. Lastly, we provide integer and float terminals to parameterize the various operators, such as the number of neighbors (k) in the k-Nearest Neighbors Classifier.

\subsection{Constructing Tree-based Pipelines}
\label{sec:constructing-tbp}

To combine these operators into a machine learning pipeline, we treat them as GP primitives and construct GP trees from them. Figure~\ref{fig:tpot-pipeline-example} shows an example tree-based pipeline, where two copies of the data set are provided to the pipeline, modified in a successive manner by each operator, combined into a single data set, and finally used to make classifications. Because all operators receive a data set as input and return the modified data set as output, it is possible to construct arbitrarily large machine learning pipelines that can act on multiple copies of the data set. Thus, GP trees provide an inherently flexible representation of machine learning pipelines.

In order for these tree-based pipelines to operate, we store three additional variables for each record in the data set. The ``class'' variable indicates the true label for each record, and is used when evaluating the accuracy of each pipeline. The ``guess'' variable indicates the pipeline's latest guess for each record, where the classifications from the last classification operator in the pipeline are stored as the ``guess''. Finally, the ``group'' variable indicates whether the record is to be used as a part of the internal training or testing set, such that the tree-based pipelines are only trained on the training data and evaluated on the testing data. We note that the data set provided to TPOT is split into an internal stratified 75\%/25\% training/testing set.

\subsection{Optimizing Tree-based Pipelines}
\label{sec:optimizing-tbp}

To automatically generate and optimize these tree-based pipelines, we use a GP algorithm~\cite{Koza1992,Banzhaf1998} as implemented in the Python package DEAP~\cite{DEAP}. The TPOT GP algorithm follows a standard GP process with settings listed in Table~\ref{table:gp-settings}. To begin, the GP algorithm generates 100 random tree-based pipelines and evaluates their accuracy on the data set. For every generation of the GP algorithm, the algorithm selects the top 20 pipelines in the population according to the NSGA-II selection scheme~\cite{Deb2002}, where pipelines are selected to simultaneously maximize classification accuracy on the data set while minimizing the number of operators in the pipeline. Each of the top 20 selected pipelines produce five offspring into the next generation's population, 5\% of those offspring experience crossover with another offspring, then 90\% of the remaining unaffected offspring experience random mutations. Every generation, the algorithm updates a Pareto front of the non-dominated solutions~\cite{Deb2002} discovered at any point in the GP run. The algorithm repeats this evaluate-select-crossover-mutate process for 100 generations---adding and tuning pipeline operators that improve classification accuracy and pruning operators that degrade classification accuracy---at which point the algorithm selects the highest-accuracy pipeline from the Pareto front as the representative ``best'' pipeline from the run.

\begin{table}[t]
    \caption{Genetic programming algorithm settings.}
    \begin{tabular}{l l}
        \hline \hline
        {\bf GP Parameter} & {\bf Value}\\ \hline
        Population size & 100\\
        Generations & 100\\
        Multi-objective selection & 5 copies of top 20\% according to NSGA-II\\
        Per-individual crossover rate & 5\%\\
        Per-individual mutation rate & 90\%\\
        Crossover & One-point crossover\\
        Mutation & Point, insert, \& shrink 1/3 chance of each\\
        Replicate runs with unique RNG seeds & 30\\
        \hline
    \end{tabular}
    \label{table:gp-settings}
\end{table}

\subsection{Sensible Initialization in TPOT}
\label{sec:sensible-initialization}

Next, we implement a version of TPOT with sensible initialization, which we call {\em TPOT-SI}. In TPOT-SI, the GP algorithm creates the initial population by seeding it with a random selection from 100 building blocks that we identified from previous TPOT runs. These building blocks consist of tree-based pipelines with 1--3 operators, e.g., one building block that we identified was ``PolynomialFeatures $\rightarrow$ LogisticRegression,'' where the building block casts the data set into a polynomial feature space then provides those features to a logistic regression model to make the classification.

The primary idea behind providing sensible initialization via building blocks is that genetic programming algorithms typically rely heavily on crossover~\cite{Poli2008}. Thus, we want to initialize the GP population with building blocks that 1) start the GP population off with pipelines that already effectively solve at least part of the classification task, and 2) can be mixed and matched to build better pipelines in a more efficient manner. However, we note that our preliminary investigations showed that a high crossover regime (per-individual crossover rate = 90\%, per-individual mutation rate = 5\%) performed worse than a high mutation regime (per-individual crossover rate = 5\%, per-individual mutation rate = 90\%) for both TPOT and TPOT-SI, so we focused on the high mutation regime in this chapter.

To identify these building blocks, we ran 30 replicates of TPOT on the 160 supervised classification benchmark data sets described in Section~\ref{sec:benchmark-data}. We identified the highest-accuracy tree-based pipeline from the final Pareto front for each run, then performed an ``n-gram'' analysis (up to $n=4$) of the pipelines to count the most frequent combinations of pipeline operators. For example, in Figure~\ref{fig:tpot-pipeline-example} ``SelectKBest $\rightarrow$ KNearestNeighborClassifier'' would be a 2-gram, and ``KNearestNeighborClassifier'' would be a 1-gram. After counting all of the n-grams in each tree-based pipeline, we summed the counts across all 4,800 replicates to determine the 100 most frequent n-grams, which we used as TPOT building blocks\footnote{Full list of building blocks:~\url{https://gist.github.com/rhiever/27f795b00b95751ee38fd9e946c72b0b}}. We have listed the top 10 most frequent building blocks in Table~\ref{table:top-building-blocks}.

\begin{table}
    \caption{10 most frequent machine learning pipeline building blocks.}
    \begin{tabular}{l}
        \hline \hline
        RandomForest\\
        XGBClassifier\\
        LogisticRegression\\
        DecisionTree\\
        KNearestNeighborClassifier\\
        XGBClassifier $\rightarrow$ RandomForest\\
        LogisticRegression $\rightarrow$ RandomForest\\
        PolynomialFeatures $\rightarrow$ LogisticRegression\\
        PolynomialFeatures $\rightarrow$ RandomForest\\
        SelectPercentile $\rightarrow$ RandomForest\\
        \hline
    \end{tabular}
    \label{table:top-building-blocks}
\end{table}

\subsection{Benchmark Data}
\label{sec:benchmark-data}

We compiled 160 supervised classification benchmarks\footnote{Benchmark data available at \url{http://www.randalolson.com/data/benchmarks/}} from a wide variety of sources, including the UCI machine learning repository~\cite{Lichman2013}, a large preexisting benchmark repository from~\cite{Reif2012}, and simulated genetic analysis data sets from~\cite{Urbanowicz2012}. These benchmark data sets range from 60 to 60,000 records, few to hundreds of features, and include binary as well as multi-class supervised classification problems. We selected data sets from a wide range of application domains, including genetic analysis, image classification, time series analysis, and many more. Thus, this benchmark represents a comprehensive suite of tests with which to evaluate automated machine learning systems.

\begin{figure}[b]
\sidecaption
\includegraphics[width=\textwidth]{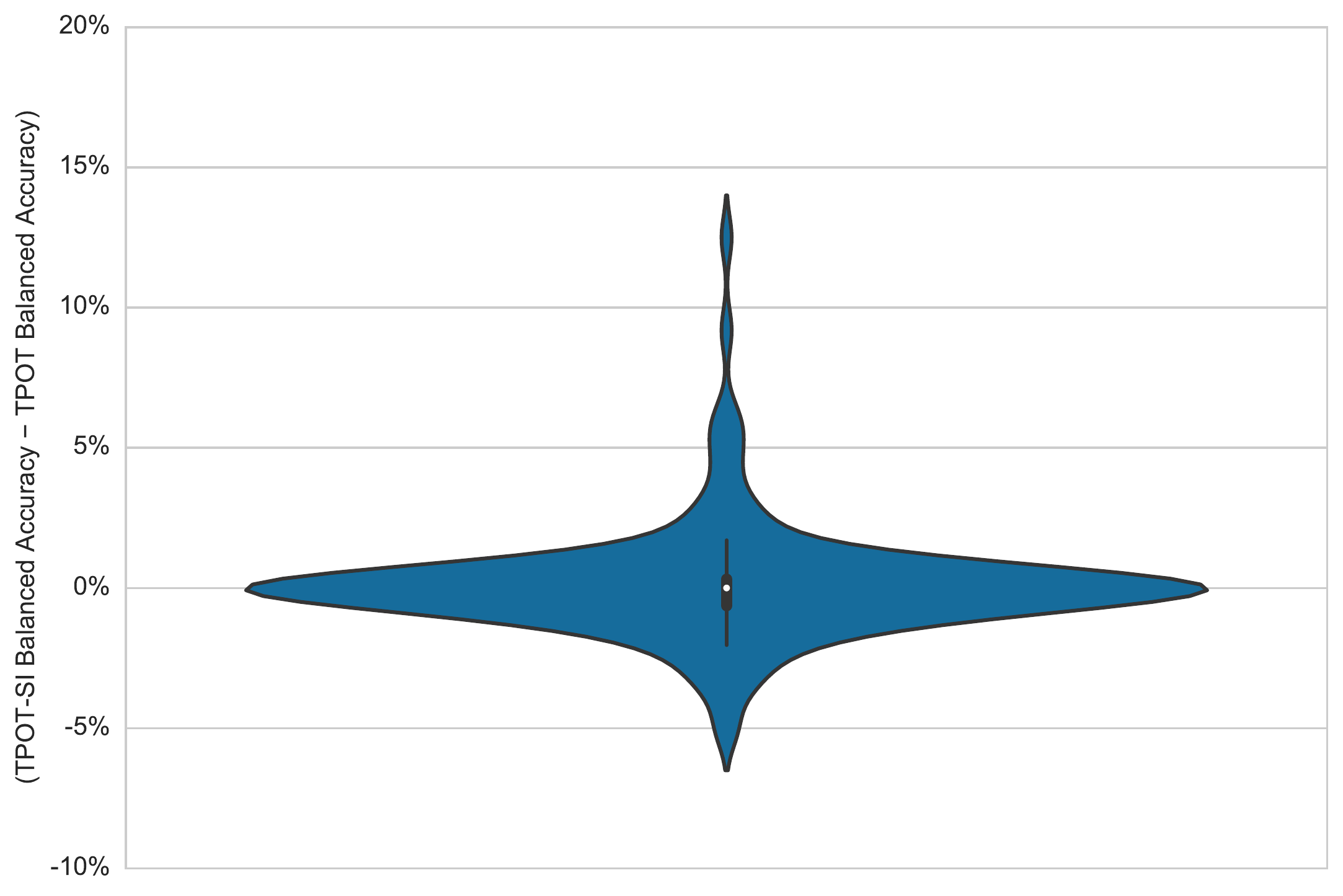}
\caption{Violin plot of the difference in median balanced accuracy between TPOT-SI and TPOT on the benchmarks. Positive values indicate an improvement in accuracy from TPOT-SI, whereas negative values indicate a degradation of accuracy from TPOT-SI. The width of the violin represents the relative density of points at that value, e.g., most differences are centered around 0\% accuracy improvement. We note that the density is estimated from the underlying data, which is why it appears that there are differences in accuracy below -5\%.}
\label{fig:tpot-vs-tpot-si-difference-violinplot}
\end{figure}

\section{Results}
\label{sec:results}

To provide an initial evaluation of TPOT-SI, we ran 30 replicates of TPOT-SI on the 160 supervised classification benchmark data sets described in Section~\ref{sec:benchmark-data}. We compare these experiments to the same experiments with a version of TPOT without sensible initialization. In all cases, we measured pipeline accuracy as balanced accuracy~\cite{Velez2007}, which corrects for class frequency imbalances in data sets by computing the accuracy on a per-class basis then averaging the per-class accuracies.

Figure~\ref{fig:tpot-vs-tpot-si-difference-violinplot} summarizes the difference in performance between TPOT-SI and TPOT on each benchmark. Overall, TPOT-SI showed no improvement on a large portion of the 160 benchmarks, small performance degradation on a few benchmarks, and fair improvement on a handful of benchmarks. Notably, the largest performance degradation was on the ``tutorial'' benchmark from~\cite{Reif2012} with a 5\% median accuracy decrease, and the largest performance increase was on the ``parity5'' benchmark from~\cite{Reif2012} with a 12.5\% median accuracy increase.

\begin{figure}[t]
\sidecaption
\includegraphics[width=\textwidth]{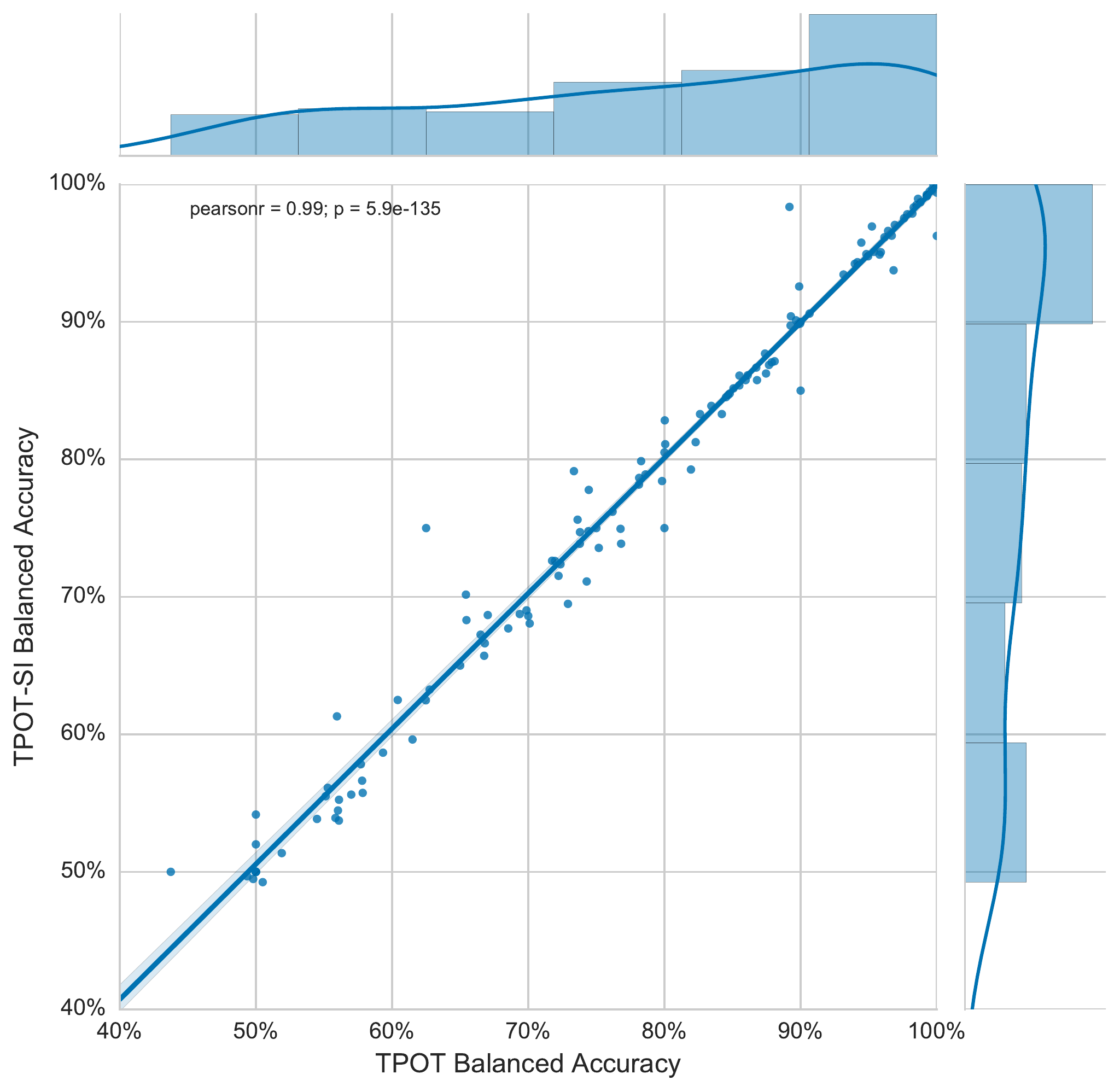}
\caption{Median balanced accuracy of TPOT-SI vs. the same for TPOT on the benchmarks. Each point represents the median balanced accuracies for one benchmark. The line represents a linear regression fit to the median accuracies, whereas the histograms on the sides show the density of the points on both axes.}
\label{fig:tpot-ba-vs-tpot-si-ba-scatter}
\end{figure}

In order to provide better insight into why so many benchmarks saw no improvement with TPOT-SI, we plotted the original TPOT accuracy on each data set vs. the TPOT-SI accuracy on the corresponding data set in Figure~\ref{fig:tpot-ba-vs-tpot-si-ba-scatter}. 49 of the benchmarks were already solved ($>90\%$ median balanced accuracy) with the base version of TPOT, which is why TPOT-SI saw no improvement on many of the benchmarks.

\begin{figure}[t]
\sidecaption
\includegraphics[width=\textwidth]{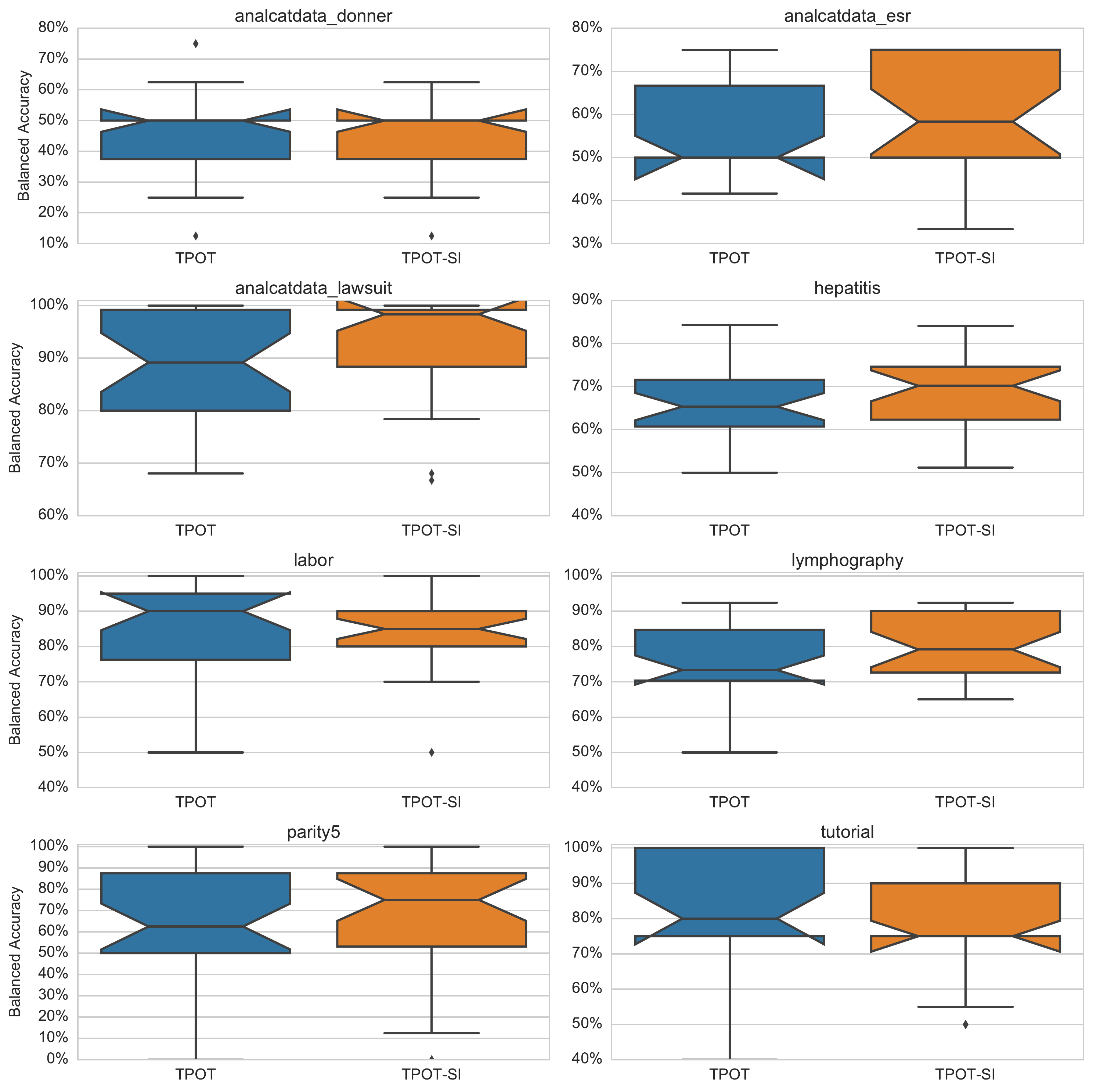}
\caption{Box plots showing the distribution of balanced accuracies for the 8 benchmarks with the biggest difference in median accuracy between TPOT-SI and TPOT. Each box plot represents 30 replicates, the inner line shows the median, and the notches represent the bootstrapped 95\% confidence interval of the median.}
\label{fig:tpot-vs-tpot-si-performance-boxplots}
\end{figure}

Finally, we show the distribution of balanced accuracies on the 8 benchmarks with the largest performance differences in Figure~\ref{fig:tpot-vs-tpot-si-performance-boxplots}. Surprisingly, the only benchmark (out of all 160) with a statistically significant difference in performance is the ``analcatdata\_lawsuit'' benchmark, where TPOT-SI allowed for a 9.2\% higher accuracy on average. In the other benchmarks, TPOT-SI allowed for small but statistically insignificant improvements over TPOT.

\section{Summary and Discussion}
\label{sec:discussion}

In this chapter, we presented preliminary results from implementing sensible initialization in TPOT. In summary, TPOT-SI saw significant improvement in performance on only one benchmark out of 160 (Figure~\ref{fig:tpot-vs-tpot-si-performance-boxplots}). Although our sensible initialization method could likely use improvement, we note that TPOT-SI did not significantly degrade performance on any of the benchmarks as well.

Of course, the key goal of this chapter extends beyond implementing a sensible initialization method in TPOT: We seek to identify the building blocks of machine learning pipelines, which is information that can be harnessed in many machine learning applications. In this chapter, we suggest that machine learning building blocks are small sequences of machine learning operators that occur frequently in pipelines used to solve benchmark classification problems. Although the building blocks that we identified do not seem to significantly improve performance via sensible initialization on many benchmarks, these results could be for many reasons. For example, that not all of the building blocks that we used are useful for all of the benchmarks, and in fact some could be detrimental on some benchmarks. We discuss how these limitations can be overcome in Section~\ref{sec:looking-forward}.

Furthermore, TPOT-SI automatically optimized pipelines for all 160 benchmarks, and discovered pipelines that achieve $>90\%$ median balanced accuracy on 52 of the benchmarks (and $>80\%$ on 79 of them) without any prior knowledge of the problem domains. These results show significant promise for GP-based automated machine learning systems. We note, however, that TPOT should not be considered a replacement for machine learning practitioners; rather, TPOT saves practitioners time by automating the tedious portions of machine learning pipeline design, but ultimately the practitioners will be responsible for deciding on the final pipeline. Similarly, we consider TPOT to be a ``Data Science Assistant'' and idea generator because it can discover unique ways to model data sets---and export its finding to the corresponding Python code---so practitioners can take TPOT pipelines and customize them for their particular application. To aid in the effort of providing an easily accessible GP-based automated machine learning system, we have released TPOT as an open source Python application at~\url{https://github.com/rhiever/tpot}.

\section{Looking Forward}
\label{sec:looking-forward}

The sensible initialization method implemented in TPOT-SI is quite simple, and there are many refinements that can be made. For example, we can use meta-learning techniques to intelligently match building blocks and pipeline configurations that will work well on the particular data set being analyzed~\cite{Feurer2015Initializing}. In short, meta-learning uses information from previous machine learning runs to estimate how well each pipeline configuration will work on a particular data set. To place data sets on a standard scale, meta-learners compute meta-features from data sets, such as data set size, the number of features, and various aspects about the features, which are then used to map data set meta-features to corresponding pipeline configurations that work well on data sets with those meta-features. Such an intelligent meta-learning algorithm is likely to improve the TPOT sensible initialization process.

Similarly, we can harness expert knowledge about machine learning building blocks to bias the GP mutation and crossover operations, similar to what was done in~\cite{Greene2007}. In this case, we would provide the GP algorithm information about how well particular pipeline combinations perform on average, e.g., ``Replacing a RandomForest operator with a DecisionTree operator is 89\% likely to degrade accuracy.'' This information could then be used to bias the mutation and crossover operations toward producing better pipelines. Further, this information could be learned and refined over the optimization process, such that the GP algorithm will learn what makes an effective pipeline for the particular data set being analyzed.

Finally, population-based optimization methods such as GP are typically criticized for maintaining a large population of solutions, which can prove to be slow and wasteful for certain optimization problems. In this case, we can turn GP's purported weakness into a strength by creating an ensemble out of the GP populations.~\cite{Bhowan2013} explored this population ensemble method previously with standard GP and showed significant improvement, and it is a natural extension to create ensembles out of TPOT's population of machine learning pipelines.

In conclusion, automated machine learning is a field of research that is ripe for GP systems. We should focus our efforts on refining a GP-based automated machine learning system, and in particular highlight GP's strengths as compared to Bayesian optimization, simulated annealing, and greedy optimization techniques. TPOT represents our effort toward this goal, and we will continue to refine TPOT until it consistently produces human-competitive machine learning pipelines.

\begin{acknowledgement}
We thank the Penn Medicine Academic Computing Services for the use of their computing resources. This work was supported by National Institutes of Health grants LM009012, LM010098, and EY022300.
\end{acknowledgement}

\bibliographystyle{spmpsci}
\bibliography{olson}

\begin{thebibliography}{10}
\providecommand{\url}[1]{{#1}}
\providecommand{\urlprefix}{URL }
\expandafter\ifx\csname urlstyle\endcsname\relax
  \providecommand{\doi}[1]{DOI~\discretionary{}{}{}#1}\else
  \providecommand{\doi}{DOI~\discretionary{}{}{}\begingroup
  \urlstyle{rm}\Url}\fi

\bibitem{Banzhaf1998}
Banzhaf, W., Nordin, P., Keller, R.E., Francone, F.D.: {Genetic Programming: An
  Introduction}.
\newblock Morgan Kaufmann, San Meateo, CA, USA (1998)

\bibitem{Bergstra2012}
Bergstra, J., Bengio, Y.: {Random Search for Hyper-Parameter Optimization}.
\newblock Journal of Machine Learning Research \textbf{13}, 281--305 (2012)

\bibitem{Bhowan2013}
Bhowan, U., Johnston, M., Zhang, M., Yao, X.: Evolving diverse ensembles using
  genetic programming for classification with unbalanced data.
\newblock Trans. Evol. Comp \textbf{17}(3), 368--386 (2013)

\bibitem{Chen2016}
Chen, T., Guestrin, C.: {XGBoost: A Scalable Tree Boosting System}.
\newblock CoRR \textbf{abs/1603.02754} (2016).
\newblock \urlprefix\url{http://arxiv.org/abs/1603.02754}

\bibitem{Deb2002}
Deb, K., Pratap, A., Agarwal, S., Meyarivan, T.: {A fast and elitist
  multiobjective genetic algorithm: NSGA-II}.
\newblock {IEEE Transactions on Evolutionary Computation} \textbf{6}, 182--197
  (2002)

\bibitem{Fuerer2015}
Feurer, M., Klein, A., Eggensperger, K., Springenberg, J., Blum, M., Hutter,
  F.: Efficient and robust automated machine learning.
\newblock In: C.~Cortes, N.~Lawrence, D.~Lee, M.~Sugiyama, R.~Garnett (eds.)
  Advances in Neural Information Processing Systems 28, pp. 2944--2952. Curran
  Associates, Inc. (2015)

\bibitem{Feurer2015Initializing}
Feurer, M., Springenberg, J.T., Hutter, F.: Initializing bayesian
  hyperparameter optimization via meta-learning.
\newblock In: Proceedings of the 29th {AAAI} Conference on Artificial
  Intelligence, January 25-30, 2015, Austin, Texas, {USA.}, pp. 1128--1135
  (2015)

\bibitem{DEAP}
Fortin, F.A., {De Rainville}, F.M., Gardner, M.A., Parizeau, M., Gagn\'e, C.:
  {DEAP: Evolutionary Algorithms Made Easy}.
\newblock Journal of Machine Learning Research \textbf{13}, 2171--2175 (2012)

\bibitem{GarciaArnau2007}
García-Arnau, M., Manrique, D., Ríos, J., Rodríguez-Patón, A.:
  Initialization method for grammar-guided genetic programming.
\newblock Knowledge-Based Systems \textbf{20}, 127--133 (2007).
\newblock The 26th {SGAI} International Conference on Innovative Techniques and
  Applications of Artificial Intelligence

\bibitem{Goldberg2002}
Goldberg, D.E.: {The Design of Innovation: Lessons from and for Competent
  Genetic Algorithms}.
\newblock Kluwer Academic Publishers, Norwell, MA, USA (2002)

\bibitem{Greene2007}
Greene, C.S., White, B.C., Moore, J.H.: An expert knowledge-guided mutation
  operator for genome-wide genetic analysis using genetic programming.
\newblock In: Pattern Recognition in Bioinformatics, pp. 30--40. Springer
  Berlin Heidelberg (2007)

\bibitem{Greene2009}
Greene, C.S., White, B.C., Moore, J.H.: Sensible initialization using expert
  knowledge for genome-wide analysis of epistasis using genetic programming.
\newblock In: 2009 IEEE Congress on Evolutionary Computation, pp. 1289--1296
  (2009)

\bibitem{MachineLearningBook}
Hastie, T.J., Tibshirani, R.J., Friedman, J.H.: The Elements of Statistical
  Learning: Data Mining, Inference, and Prediction.
\newblock Springer, New York, NY, USA (2009)

\bibitem{Hutter2015}
Hutter, F., L{\"u}cke, J., Schmidt-Thieme, L.: {Beyond Manual Tuning of
  Hyperparameters}.
\newblock KI - K{\"u}nstliche Intelligenz \textbf{29}, 329--337 (2015)

\bibitem{Kanter2015}
Kanter, J.M., Veeramachaneni, K.: {Deep Feature Synthesis: Towards Automating
  Data Science Endeavors}.
\newblock In: Proceedings of the International Conference on Data Science and
  Advance Analytics. IEEE (2015)

\bibitem{Koza1992}
Koza, J.R.: Genetic Programming: On the Programming of Computers by Means of
  Natural Selection.
\newblock MIT Press, Cambridge, MA, USA (1992)

\bibitem{Lichman2013}
Lichman, M.: {UCI} machine learning repository (2013).
\newblock {http://archive.ics.uci.edu/ml}

\bibitem{Luke01}
Luke, S., Panait, L.: A survey and comparison of tree generation algorithms.
\newblock In: L.~Spector, E.D. Goodman, A.~Wu, W.B. Langdon, H.M. Voigt,
  M.~Gen, S.~Sen, M.~Dorigo, S.~Pezeshk, M.H. Garzon, E.~Burke (eds.)
  Proceedings of the 6th Genetic and Evolutionary Computation Conference, GECCO
  '01, pp. 81--88. Morgan Kaufmann, San Francisco, California, USA (2001)

\bibitem{Martinsson2011}
Martinsson, P.G., Rokhlin, V., Tygert, M.: A randomized algorithm for the
  decomposition of matrices.
\newblock Applied and Computational Harmonic Analysis \textbf{30}, 47--68
  (2011)

\bibitem{Olson2016GECCO}
Olson, R.S., Bartley, N., Urbanowicz, R.J., Moore, J.H.: Evaluation of a
  tree-based pipeline optimization tool for automating data science.
\newblock arXiv e-print. http://arxiv.org/abs/1603.06212 (2016)

\bibitem{Olson2016EvoBio}
Olson, R.S., Urbanowicz, R.J., Andrews, P.C., Lavender, N.A., Kidd, L.C.,
  Moore, J.H.: Applications of Evolutionary Computation: 19th European
  Conference, EvoApplications 2016, Porto, Portugal, March 30 — April 1,
  2016, Proceedings, Part I, chap. Automating Biomedical Data Science Through
  Tree-Based Pipeline Optimization, pp. 123--137.
\newblock Springer International Publishing (2016)

\bibitem{ONeill2003}
O'Neill, M., Ryan, C.: Grammatical Evolution: Evolutionary Automatic
  Programming in a Arbitrary Language, \emph{Genetic programming}, vol.~4.
\newblock Kluwer Academic Publishers (2003)

\bibitem{scikit-learn}
Pedregosa, F., Varoquaux, G., Gramfort, A., Michel, V., Thirion, B., Grisel,
  O., Blondel, M., Prettenhofer, P., Weiss, R., Dubourg, V., Vanderplas, J.,
  Passos, A., Cournapeau, D., Brucher, M., Perrot, M., Duchesnay, E.:
  Scikit-learn: Machine learning in {P}ython.
\newblock Journal of Machine Learning Research \textbf{12}, 2825--2830 (2011)

\bibitem{Poli2008}
Poli, R., Langdon, W.B., McPhee, N.F.: A Field Guide to Genetic Programming.
\newblock Lulu Enterprises, UK Ltd (2008)

\bibitem{Reif2012}
Reif, M.: A comprehensive dataset for evaluating approaches of various
  meta-learning tasks.
\newblock In: First International Conference on Pattern Recognition and Methods
  (ICPRAM), First International Conference on Pattern Recognition and Methods
  (ICPRAM) (2012)

\bibitem{Simon2013}
Simon, P.: Too big to ignore: the business case for big data.
\newblock Wiley \& SAS Business Series. Wiley, New Delhi (2013)

\bibitem{Snoek2012}
Snoek, J., Larochelle, H., Adams, R.P.: {Practical Bayesian Optimization of
  Machine Learning Algorithms}.
\newblock In: F.~Pereira, C.J.C. Burges, L.~Bottou, K.Q. Weinberger (eds.)
  Advances in Neural Information Processing Systems 25, pp. 2951--2959. Curran
  Associates, Inc. (2012)

\bibitem{Urbanowicz2012}
Urbanowicz, R.J., Kiralis, J., Sinnott-Armstrong, N.A., Heberling, T., Fisher,
  J.M., Moore, J.H.: {GAMETES: a fast, direct algorithm for generating pure,
  strict, epistatic models with random architectures}.
\newblock {BioData Mining} \textbf{5} (2012)

\bibitem{Velez2007}
Velez, D.R., et~al.: {A balanced accuracy function for epistasis modeling in
  imbalanced datasets using multifactor dimensionality reduction}.
\newblock Genetic Epidemiology \textbf{31}(4), 306--315 (2007)

\bibitem{Zutty2015}
Zutty, J., Long, D., Adams, H., Bennett, G., Baxter, C.: Multiple objective
  vector-based genetic programming using human-derived primitives.
\newblock In: Proceedings of the 2015 Annual Conference on Genetic and
  Evolutionary Computation, GECCO '15, pp. 1127--1134. ACM, New York, NY, USA
  (2015)

\end{thebibliography}

\end{document}